\documentclass[letterpaper, 10 pt, conference]{ieeeconf}  

\IEEEoverridecommandlockouts 

\usepackage[utf8]{inputenc} 
\usepackage[T1]{fontenc}    
\usepackage{hyperref}       
\usepackage{url}            
\usepackage{wrapfig,booktabs}       
\usepackage{amsfonts}       
\usepackage{nicefrac}       
\usepackage{microtype}      
\usepackage{xcolor}         
\usepackage{amsmath}
\usepackage{amssymb}
\usepackage{pgf-pie}
\usepackage{graphicx}
\usepackage{pgfplots}
\usetikzlibrary{patterns}
\pgfplotsset{compat=1.18}
\usepackage{graphicx}
\usepackage{pythonhighlight}
\usepackage{bbding}

\newcommand{\datasetname}{\textsc{Can-Do}}
\newcommand{\methodname}{NeuroGround}
\newcommand{\generatorname}{DALL-E 3}
\newcommand{\datasize}{400}

\title{Can-Do! A Dataset and Grounded Prompting for Embodied Planning with Large Multimodal Models}

\title{Can-Do! A Dataset and Neuro-Symbolic Grounded Prompting for Embodied Planning with Large Multimodal Models}

\title{\textit{I Can, Therefore I Do}\\Neuro-Symbolic Grounded Embodied Planning with Large Multimodal Models}

\title{Can-Do! A Dataset and Neuro-Symbolic Grounded Framework for \\ Embodied Planning with Large Multimodal Models}

\author{
Yew Ken Chia$^{1,2 *}$,
Qi Sun$^{1 *}$,
Lidong Bing$^{2}$,
Soujanya Poria$^{1}$ \\
$^1$ Singapore University of Technology and Design \quad $^2$ DAMO Academy, Alibaba Group 
\thanks{$^{*}$Both authors contributed equally. Yew Ken Chia is under the Joint Ph.D. Program between DAMO Academy and the Singapore University of Technology and Design.} 
}

\begin{document}

\maketitle

\begin{abstract}
Large multimodal models have demonstrated impressive problem-solving abilities in vision and language tasks, and have the potential to encode extensive world knowledge. However, it remains an open challenge for these models to perceive, reason, plan, and act in realistic environments. In this work, we introduce Can-Do, a benchmark dataset designed to evaluate embodied planning abilities through more diverse and complex scenarios than previous datasets. Our dataset includes \datasize{} multimodal samples, each consisting of natural language user instructions, visual images depicting the environment, state changes, and corresponding action plans. The data encompasses diverse aspects of commonsense knowledge, physical understanding, and safety awareness. Our fine-grained analysis reveals that state-of-the-art models, including GPT-4V, face bottlenecks in visual perception, comprehension, and reasoning abilities. To address these challenges, we propose \methodname{}, a neuro-symbolic framework that first grounds the plan generation in the perceived environment states and then leverages symbolic planning engines to augment the model-generated plans. Experimental results demonstrate the effectiveness of our framework compared to strong baselines. Our code and dataset are available at \url{https://embodied-planning.github.io}.
\end{abstract}

\section{Introduction}

Large multimodal models extend the general capabilities of large language models by integrating aspects such as visual understanding \cite{dawn}.
Impressively, they demonstrate the potential for real-world problem solving through reasoning and acting on multimodal inputs \cite{yang2023mmreact}
However, it remains an open challenge for state-of-the-art models to effectively perceive, reason, plan, and act in complex practical situations \cite{saycan, vila}.
Concretely, we envision the future of large multimodal models as capable embodied planners \cite{huang2022innermonologue}.

However, a potential bottleneck in advancing large multimodal models is the lack of suitable embodied planning benchmarks. 
While existing benchmarks such as ALFRED \cite{ALFRED20} and TaPA \cite{TaPA} do support multimodal scenarios, their rigidly defined environments may hinder the development of flexible and diverse planning scenarios.
On the other hand, we observe that majority of the existing data focus on short to medium-length plans and explicit user instructions, which may not provide sufficient complexity for the rapidly advancing models.
Therefore, in this work, we introduce \datasetname{}, an embodied planning dataset for benchmarking large multimodal models.
As shown in Table \ref{tab:data_comparision_appendix}, our dataset focuses on what we envision that effective embodied planners can do: To generalize to realistic and diverse scenarios in multimodal environments, tackling complex or ambiguous planning problems.


\begin{table}[!t]
\centering
\caption{Comparison of benchmarks for embodied task planning.}
\vspace{-0.6em}
\resizebox{\linewidth}{!}{
\begin{tabular}{lcccc}
\toprule
 & ALFRED & TaPA & \textbf{\datasetname{} (Ours)}  \\
\midrule
Synthetic Images & \color{teal}{\Checkmark} & \color{teal}{\Checkmark} & \color{teal}{\Checkmark} \\
Real Images & \color{purple}{\XSolidBrush} & \color{purple}{\XSolidBrush} & \color{teal}{\Checkmark} \\
Commonsense Reasoning & \color{purple}{\XSolidBrush} & \color{teal}{\Checkmark} & \color{teal}{\Checkmark} \\
Physical Understanding & \color{purple}{\XSolidBrush} & \color{purple}{\XSolidBrush} & \color{teal}{\Checkmark} \\
Safety \& Hazards Awareness & \color{purple}{\XSolidBrush} & \color{purple}{\XSolidBrush} & \color{teal}{\Checkmark} \\
\bottomrule
\end{tabular}
}
\vspace{-1em}
\label{tab:data_comparision_appendix}
\end{table}

Based on our fine-grained analysis of state-of-the-art models, we discover that existing models, including GPT-4V, face significant bottlenecks in visual perception, comprehension, and reasoning.
This is primarily reflected in the following aspects: 1) when a model has weak visual perception, it may be unable to perceive the state of the environment.
2) when a model has poor comprehension ability, it may misinterpret the outcome or goal state that is required to satisfy the user intent. 
3) when a model cannot reason step-by-step, it may generate an invalid plan that cannot achieve the goal.

Hence, there is an urgent need to address the bottlenecks of embodied planning with large multimodal models. To this end, we further propose \methodname{}, a neuro-symbolic framework which explicitly grounds the plan generation process in the initial and goal states of the environment, and consequently leverages symbolic engines to augment the model-generated plans.

Experiments across multiple planning categories in \datasetname{} and state-of-the-art models show that our proposed framework demonstrates significant benefits compared to strong baselines such as chain-of-thought prompting.
Thus, our main contributions can be summarized as follows: 
\begin{itemize}
\item We introduce \datasetname{}, an embodied planning dataset that can benchmark large multimodal models with more diverse and complex scenarios.
\item Based on our fine-grained analysis, we diagnose significant planning bottlenecks in state-of-the-art models: visual perception, comprehension, and reasoning.
\item To address the planning bottlenecks, we propose \methodname{}, a framework which leverages state-grounded planning and augments models with symbolic engines.
\end{itemize}

\begin{figure*}[t]
    \centering
    \includegraphics[width=0.9\textwidth]{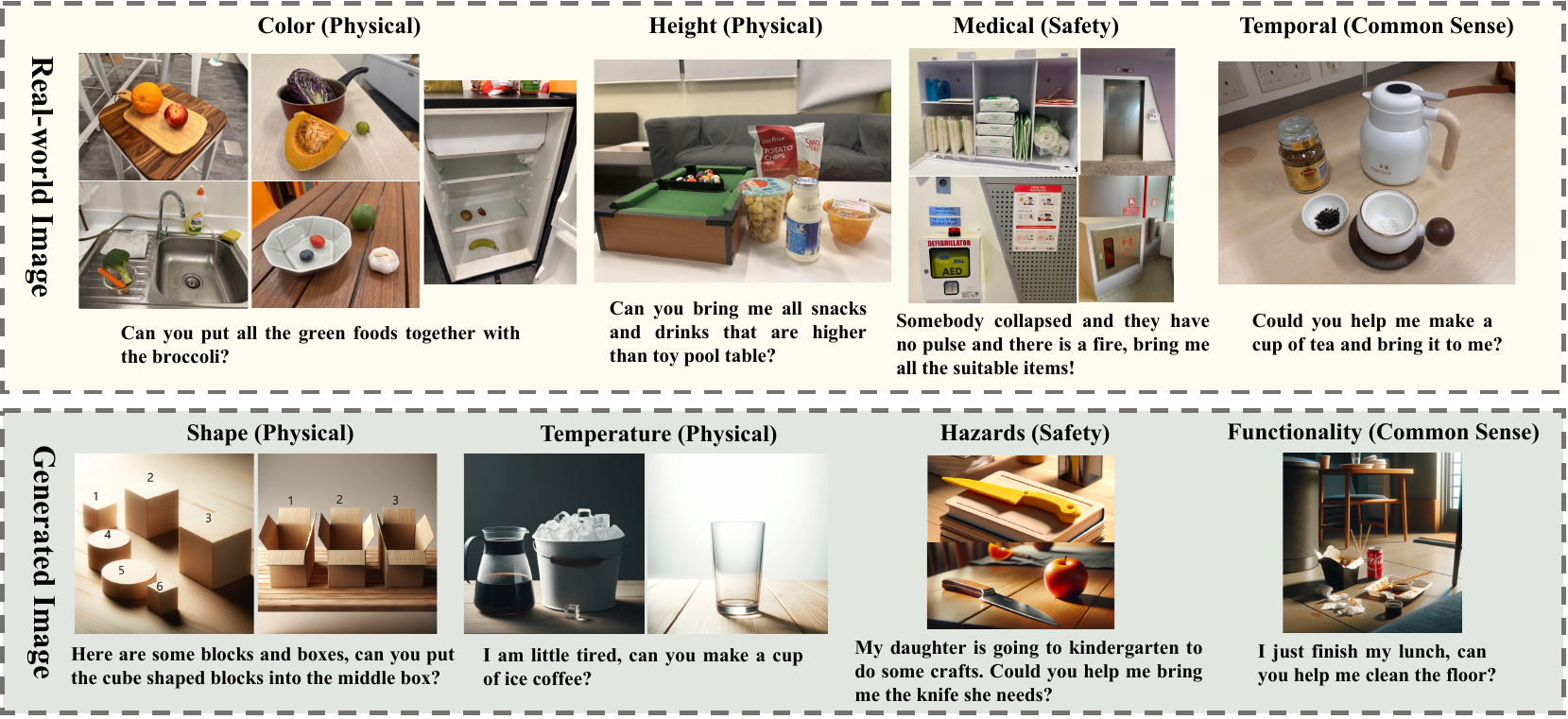}
    \vspace{-0.6em}
    \caption{Examples of embodied planning problems in our \datasetname{} dataset.}
    \label{fig:dataset_sample}
    \vspace{-1em}
\end{figure*}

\section{Related Work}

\paragraph{Large Multimodal Models}

Recently, models such as GPT-4V have gained widespread attention through impressive visual and language capabilities such as composing stories or solving math problems from real images \cite{yin2024multimodalsurvey}.
To this end, there has been rapid development to advance large multimodal models, including closed-source models such as Claude Opus and Gemini Pro, as well as open-source models like LLaVA \cite{liu2023visualllava} and Qwen-VL \cite{bai2024qwenvl}.
In this work, we mainly focus on the leading models such as GPT-4V to explore the limits of embodied planning, which are currently closed-source.

\paragraph{Embodied Planning}

The advancement of large multimodal models have also enabled new opportunities to solve planning problems at the intersection of computer vision, natural language processing, and embodied artificial intelligence \cite{embodiedsurvey}.
To foster the development models that can generalize to perceive, reason, plan, and act in the real-world \cite{hao2023reasoningrap}, we are inspired by existing datasets like ALFRED \cite{ALFRED20} and TaPA \cite{TaPA} to explore more diverse and complex planning scenarios.
Compared to prior works, our dataset \datasetname{} flexibly supports any images that visually depict the environment, enabling diverse possibilities of locations, objects, and layouts.
Furthermore, we focus on challenging user instructions which may not explicitly specify the actions, and instead require models to comprehend the user intent.

\section{\datasetname{}: A Multimodal Benchmark Dataset for Embodied Task Planning}
\label{sec:dataset}

In this section, we introduce the construction of our \datasetname{} dataset.
To explore diverse and complex scenarios for evaluating the planning abilities of the multimodal models, we leverage both real scene images and synthetic images to depict the environment in our dataset, which we found to be realistic and of high quality. Thus, our dataset supports greater flexibility to generate diverse and complex objects and scenarios.
Based on this dataset, we evaluate the ability of multimodal models to perceive visual scenarios, comprehend the user intent, and generate step-by-step plans. We provide several examples of our \datasetname{} in Figure \ref{fig:dataset_sample}.

\subsection{Data Construction}
To construct diverse embodied planning scenarios, we consider three task categories that are applicable to everyday robotic assistant scenarios \cite{wang-etal-2023-newton}: physical understanding, commonsense, and safety.
Our data collection process includes the following three steps:

\paragraph{Image Generation} While our dataset is based on real scene images, we are constrained in the diversity of images and scenarios that can be captured. 
For example, it may not be feasible or ethical to recreate safety hazard cases in real life. Thus, we additionally curated a collection of locations, objects, and scenarios inspired by real-life situations, and then prompted \generatorname{}\footnote{\url{https://openai.com/index/dall-e-3/}} to generate images of the environment.
However, the primary challenge in image generation is ensuring that the image generator can controllably produce scenes focusing on specific locations or objects without producing overly complex or exaggerated images. 
To tackle this, we employed two strategies to improve the image generation process when the generated images of the given scenario are hard to control. Firstly, we can reduce the complexity of the scene by reducing the number of objects or locations. 
Secondly, we prompt the image generator with descriptions of simple styles and bright backgrounds, thereby reducing the likelihood of generating extraneous objects.

\paragraph{Instruction and State Annotation} For each scene of images, we manually design the questions and annotate the initial and goal state according to the user instructions. Our data design principles primarily focus on two aspects. Firstly, the instruction needs to test the ability of the multimodal model in specific categories. For instance, as shown in Figure \ref{fig:dataset_sample}, the physical understanding problem requires the model to recognize the shape of blocks in the image when planning. 
Consequently, we annotate the initial and goal states, which can determine the action plan.

\begin{figure}[t!]
        \centering
        \includegraphics[width=0.24\textwidth]{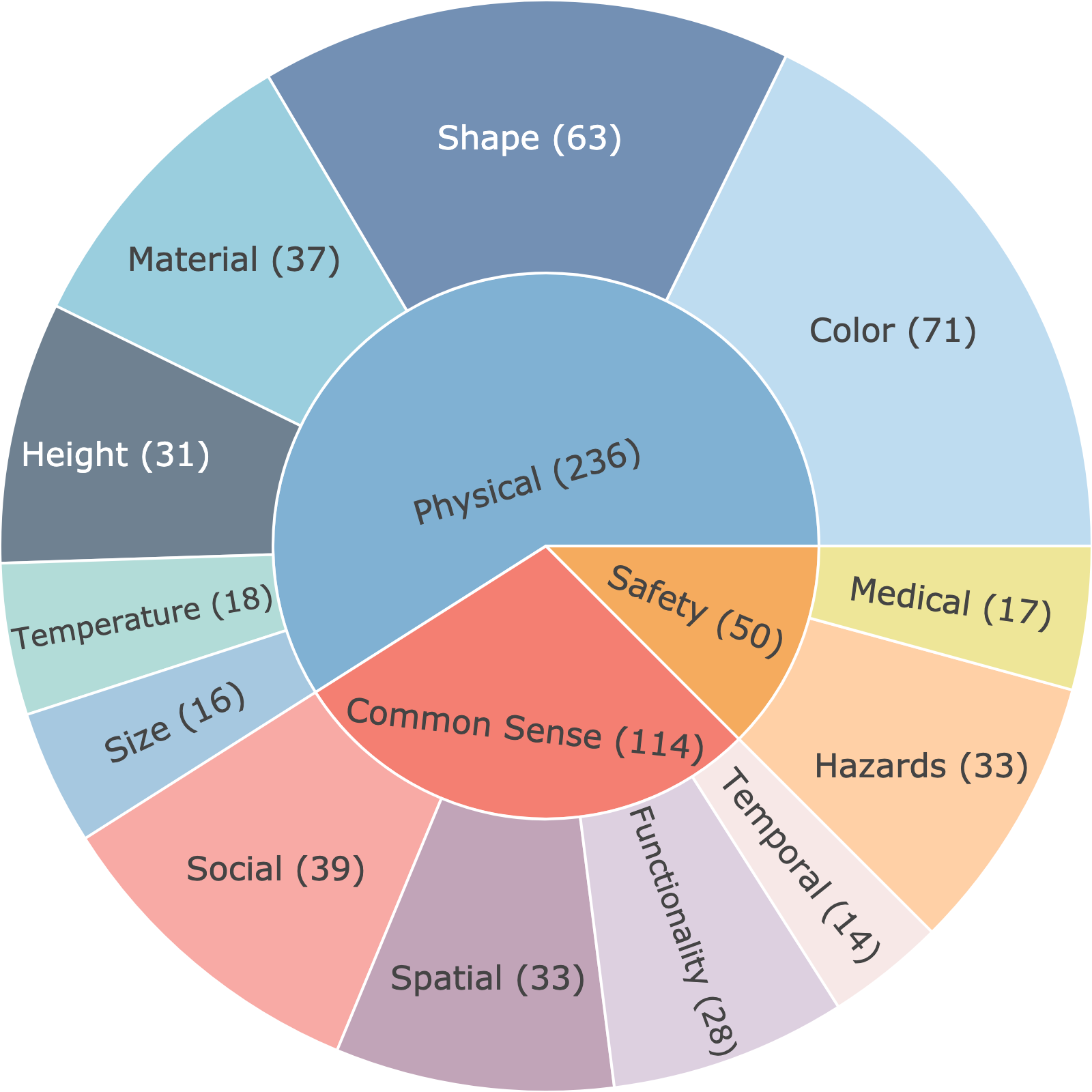}
        \vspace{-0.6em}
        \caption{Data statistics and distribution.
        }
        \vspace{-1em}
        \label{fig:data_stats}
\end{figure}%

\paragraph{Data Verification}
To ensure data quality, we engage five data annotators. Two of them are responsible for constructing the data, while the others validated the annotations.
They are also asked to flag the images for sensitive or offensive content. 
Hence, we do not expect a negative impact on society from the dataset.
We will release the dataset publicly with the CC BY-SA license.

\paragraph{Analysis}

\begin{figure}[t!]
        \centering
        \begin{tikzpicture}
    \definecolor{customRed}{HTML}{F5867F}
    \definecolor{customYellow}{HTML}{FFBC80}
    \definecolor{customBlue}{HTML}{6B98C4}
    
        \begin{axis}[
            ybar,
            bar width=.22cm,
            width=0.36\textwidth,
            height=4.6cm,
            enlarge x limits=0.15,
            ylabel={Percentage (\%)},
            xlabel={Number of Actions},
            symbolic x coords={1-4, 5-7, 8-10, 11-13, 14+},
            xtick=data,
            ymin=0,
            ymax=70,
            ytick={0,10,20,30,40,50,60},
            grid=major,
            xmajorgrids=false, 
            grid style={dashed,gray!30},
                    legend style={
                font=\fontsize{8}{1}\selectfont, 
                legend style={row sep=-0.1cm},
                at={(1,1)},
            },
            legend image code/.code={
              \draw[#1] (0cm,-0.1cm) rectangle (0.4cm,0.1cm);
            }, 
        ]
        \addplot [fill=customRed] coordinates {(1-4,55) (5-7,33) (8-10,9) (11-13,3) (14+,0)};
        \addplot [fill=customYellow] coordinates {(1-4,47) (5-7,20) (8-10,11) (11-13,19) (14+,4)};
        \addplot [fill=customBlue] coordinates {(1-4,7.8) (5-7,47.5) (8-10,7.2) (11-13,25.5) (14+,12.0)};
        \legend{ALFRED, TaPA, \datasetname{}}
        \end{axis}
    \end{tikzpicture}
    \vspace{-0.6em}
    \caption{Comparison of plan lengths of datasets. }
    \vspace{-1em}
    \label{fig:lengths}
\end{figure}
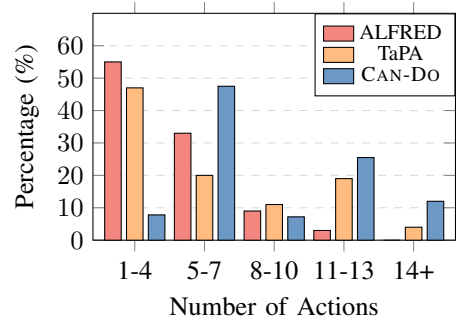

We collected \datasize{} data samples across three categories, each containing multiple aspects, as shown in Figure \ref{fig:data_stats}. 
This sample size was chosen to allow for detailed annotation and analysis within the scope of this study, supporting diverse evaluation aspects.
As our data contains a significant amount of real-image samples (18.5\%), we believe it is an important step towards bridging the gap between existing benchmarks and real-world scenarios.
As many applications use synthetic or simulated data, we believe such a hybrid approach is promising for scalable and realistic data collection.
Furthermore, in comparison to previous datasets in Figure \ref{fig:lengths}, the tasks in our dataset generally show more complexity in terms of planning lengths.



\section{Preliminary Study on Embodied Planning Bottlenecks}
\label{sec:bottlenecks}

While large multimodal models have demonstrated impressive capabilities in reasoning and problem-solving \cite{sparks}, it is not clear how this advantage extends to embodied planning tasks.
Specifically, embodied planning involves multiple stages, including the visual perception of the environment, comprehending the goal of user requests, and generating a coherent plan that satisfies the environment constraints and planning goal.
In practice, large multimodal models may face planning bottlenecks in any of these stages.
Hence, we first conduct a preliminary study to diagnose potential planning bottlenecks.
Concretely, we leverage a progressive prompting approach, by providing partial ground-truth information in the model input prompt.
For example, providing the ground-truth initial state of the environment reduces the need for the model to perceive the environment visually.
Based on the results in Section \ref{sec:prelim_results}, we find significant bottlenecks in visual perception, comprehension, and reasoning.

\subsection{Task Formulation}
\label{sec:formulation}

In this section, we aim to provide a concrete formulation of the embodied planning task. 
In general, the model inputs can contain multimodal data such as images depicting the environment and textual user instructions or task descriptions.
The main output will be the action plan, which is a sequence of actions specifying what the agent will perform.

Specifically, the environment has an initial state $S_{\text{init}}$ which represents the mapping between locations $X_{\text{loc}} = \{x_{\text{loc}_1}, \ldots, x_{\text{loc}_m}\}$ and objects $X_{\text{obj}} = \{x_{\text{obj}_1}, \ldots, x_{\text{obj}_p}\}$.
Given a textual user request $X_{\text{req}}$ and a set of visual inputs $X_{\text{img}} = \{x_{\text{img}_1}, \ldots, x_{\text{img}_n}\}$ depicting the environment, the agent $M$ should generate the action plan, a sequence of actions to fulfill the request $X_{\text{req}}$.
The plan is valid if it reaches the implicit goal state $S_{\text{goal}}$, which represents the mapping between the locations and objects that fulfill the user request. Please note that the states $S_{\text{init}}$ and $S_{\text{goal}}$ are ground-truth annotations provided in our dataset.
Note that the action types are a predefined list, included in model inputs.

To avoid ambiguity in evaluating model-generated plans, we provide the exact names of the locations $X_{\text{loc}}$ and objects $X_{\text{obj}}$, but the agent needs to perceive the visual inputs $X_{\text{img}}$ to determine the specific initial state $S_{\text{init}}$ of the environment.
The desired output is a plan consisting of a sequence of actions: $P = \{(a_1, o_1), \ldots, (a_l, o_l)\}$ where $a$ denotes the action type and $o$ denotes the corresponding location or object.
To determine whether the plan $P$ is valid, we automatically execute the actions on the initial state $S_{\text{init}}$ to obtain the outcome state $S_{\text{outcome}}$.
The plan is considered valid if the outcome state $S_{\text{outcome}}$ matches the goal state $S_{\text{goal}}$, and invalid otherwise. 
Note that the plan will be considered invalid if it also contains invalid actions. 
For example, it is invalid to pick up an object if the agent is not at the same location or to put down an object if the agent did not pick it up.

\begin{figure*}[t!]
\centering
\resizebox{0.8\textwidth}{!}{
\begin{tikzpicture}
    \definecolor{customRed_1}{HTML}{EEBEC0}
    \definecolor{customRed_2}{HTML}{F5867F}
    \definecolor{customRed_3}{HTML}{C94E65}
    \definecolor{customYellow_1}{HTML}{EFDBB9}
    \definecolor{customYellow_2}{HTML}{F1B656}
    \definecolor{customYellow_3}{HTML}{D9995B}
    \definecolor{customBlue_1}{HTML}{A3CDEA}
    \definecolor{customBlue_2}{HTML}{397FC7}
    \definecolor{customBlue_3}{HTML}{015493}
    \begin{axis}[
        width=11cm,
        height=4.8cm,
        ybar=0cm,
        bar width=7pt,
        ymax=100,
        ymin=20,
        ylabel={Plan Validity (\%)},
        xtick = {1,2,3,4,5},
        xticklabels = {Claude Opus, Gemini Pro, GPT-4V},
        xtick pos = left,
        ytick pos = left,
        ymajorgrids = true,
        ytick={0,20,40,60,80,100},
        enlarge x limits=0.3,
        grid style=dashed, 
        legend style={
            legend style={row sep=-0.1cm},
            at={(1.5,1.002)},
        },
        legend image code/.code={
          \draw[#1] (0cm,-0.1cm) rectangle (0.4cm,0.05cm);
        }, 
        legend cell align={left},
    ]
    \addplot[fill=customRed_1] coordinates {(1, 31) (2, 47) (3, 37)};
    \addlegendentry{Commonsense: Baseline};
    \addplot[fill=customRed_2] coordinates {(1, 58) (2, 53) (3, 59)};
    \addlegendentry{\quad With Guided Initial State};
    \addplot[fill=customRed_3] coordinates {(1, 57) (2, 64) (3, 67)};
    \addlegendentry{\quad With Guided Initial \& Goal };
    \addplot[fill=customYellow_1] coordinates {(1, 64) (2, 78) (3, 77)};
    \addlegendentry{Physical: Baseline};
    \addplot[fill=customYellow_2] coordinates {(1, 67) (2, 82) (3, 78)};
    \addlegendentry{\quad With Guided Initial State};
    \addplot[fill=customYellow_3] coordinates {(1, 92) (2, 94) (3, 93)};
    \addlegendentry{\quad With Guided Initial \& Goal };
    \addplot[fill=customBlue_1] coordinates {(1, 48) (2, 44) (3, 43)};
    \addlegendentry{Safety: Baseline};
    \addplot[fill=customBlue_2] coordinates {(1, 51) (2, 55) (3, 58)};
    \addlegendentry{\quad With Guided Initial State};
    \addplot[fill=customBlue_3] coordinates {(1, 76) (2, 68) (3, 64)};
    \addlegendentry{\quad With Guided Initial \& Goal };
    \end{axis}
\end{tikzpicture}
}
\vspace{-0.6em}
\caption{Preliminary study on embodied planning bottlenecks across different planning categories. To investigate visual perception, comprehension, and reasoning abilities, we progressively prompt multimodal models with ground-truth information of the initial state and goal state of the environment.}
\vspace{-1em}
\label{fig:bottleneck}
\end{figure*}
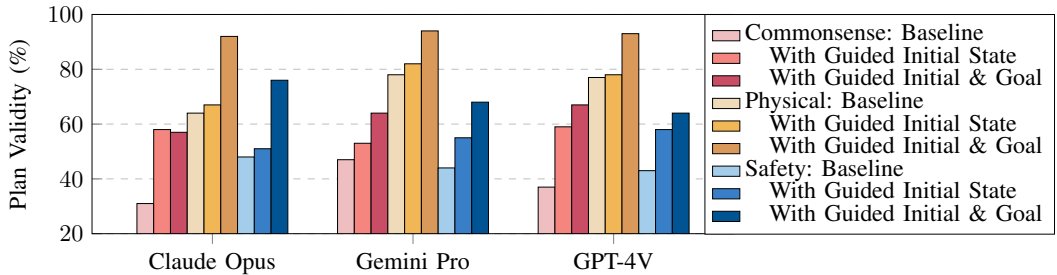

\subsection{Evaluation Setting}
\label{sec:setting}

To evaluate the performance of large multimodal models on our dataset, we provide all necessary information in the prompt, including the task description, exact location names, and object names.
To judge if a plan generated by the model is correct, we use automatic methods by executing the plan step-by-step in a symbolic environment\footnote{\url{https://unified-planning.readthedocs.io/en}}.
Concretely, the symbolic environment executes the model-generated plan and checks if the outcome state matches our ground-truth goal state, as discussed in the formulation in Section \ref{sec:formulation}.
Hence, our main evaluation metric is the average plan validity across the data samples.
To provide a more nuanced analysis, we report the average performance in each category of commonsense, physical understanding, and safety.
Similar to \textsc{SayCan} \cite{saycan}, we prompt the model to directly generate the action plan, however, we do not use a value function to rank outputs as most of the leading multimodal models do not provide the full model access to token probabilities.
While we aim to study if the models can generalize well to previously unseen environments, we find that a zero-shot evaluation setting poses many challenges in evaluating arbitrary output formats.
Hence, as language models can easily learn task formats from demonstrations \cite{gpt3}, we restrict our evaluation to the one-shot setting and provide one demonstration sample to the model based on the corresponding planning category.

\begin{table}[!t]
\centering
\caption{Analysis on alternative data settings.}
\vspace{-0.6em}
\resizebox{0.9\linewidth}{!}{
\begin{tabular}{lccccc}
\toprule
 & {Claude} & {Gemini} & {GPT-4V} \\
 \midrule
Baseline & 45.3 & 56.6 & 52.1 \\
\quad w/ Text-Only Inputs & 39.4 & 39.7 & 36.8 \\
\quad w/ Real Images Only & 29.2 & 34.3 & 35.8 \\
\bottomrule
\end{tabular}
}
\vspace{-1em}
\label{tab:analysis}
\end{table}

\subsection{Preliminary Results on Planning Bottlenecks}
\label{sec:prelim_results}


Based on the experiment results in Figure \ref{fig:bottleneck}, we observe significant planning bottlenecks when we progressively provide ground-truth state information.
In general, we find that providing the ground-truth initial state and goal state both lead to higher performance.
This indicates that the large multimodal models may face bottlenecks in the visual perception and comprehension of user instructions.
However, the performance even with state information is not close to 100\%, which indicates that reasoning to generate step-by-step plans remains a major bottleneck.
We believe that these bottlenecks should form the basis of future areas of improvement for planning with large multimodal models.

To analyze various aspects of our data, we also investigate alternative data settings in Table \ref{tab:analysis}. 
Notably, when we remove the images to form text-only model inputs, the performance greatly decreases, which demonstrates the importance of leverage visual information for our task.
To compare the challenges of samples based on real or synthetic images, we find that real image cases tend to be more challenging, which may be due to longer plan lengths of 11.7 on average compared to 9.2 for the combined data. 
Hence, the difference in performance may not be due to different image sources. 
Nevertheless, we find that the combined dataset still poses a great challenge for leading models, and we highlight the value of synthetic images for diversity and feasibility.


\begin{figure*}[t]
    \centering
    \includegraphics[width=.9\textwidth]{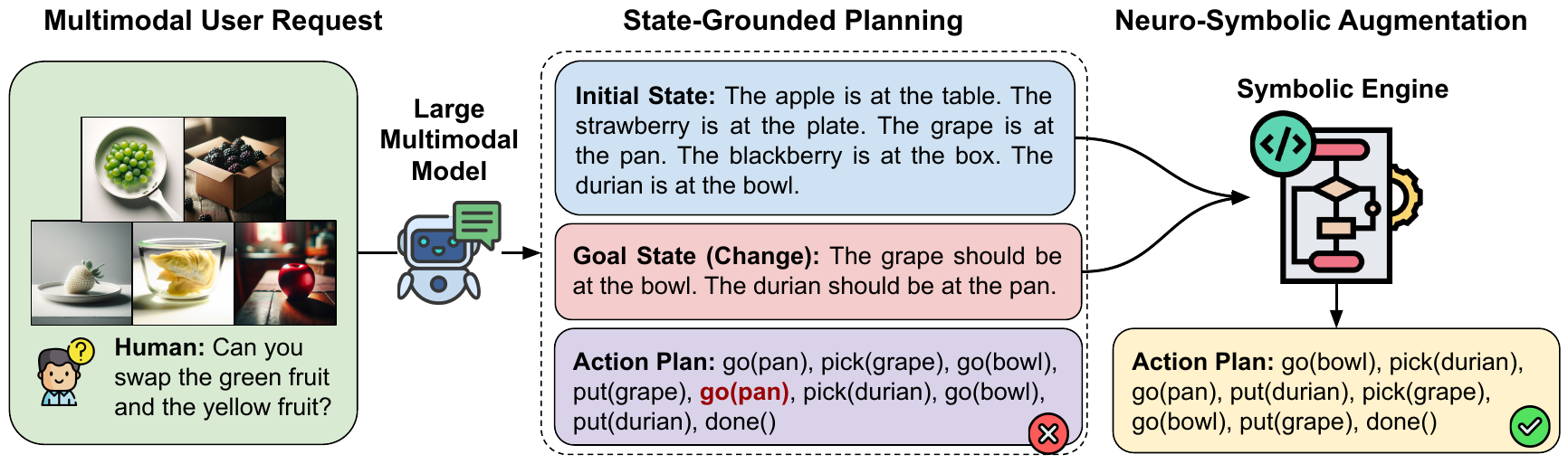}
    \vspace{-0.6em}
    \caption{An overview of \methodname{}, our neuro-symbolic framework for grounded planning. To enhance visual perception and goal comprehension of the model, we leverage state-grounded planning which explicitly guides the model to perceive and condition on the environment states before starting to generate the plan. To mitigate the plan generation challenges, we leverage a symbolic engine to augment the model-generated plan, based on the perceived states. For brevity, we do not show the task description and demonstration inputs. 
    }
    \label{fig:framework}
     \vspace{-1em}
\end{figure*}

\section{\methodname{}: A Neuro-Symbolic Framework for Grounded Planning}

Based on the preliminary study in Section \ref{sec:bottlenecks} which diagnosed significant planning bottlenecks in visual perception, goal comprehension, and reasoning for plan generation, we propose \methodname{}, a neuro-symbolic framework for grounded embodied planning.
To enhance the visual perception and comprehension ability of the model, we leverage state-grounded planning which
explicitly guides the model to generate and condition on the environment states before starting to
generate the plan. 
To mitigate the plan generation limitations of the model, we augment it with a symbolic engine, inspired by neuro-symbolic approaches \cite{shirai2024vilain, Garcez2020NeurosymbolicAT}.
In this way, our framework is able to enhance the planning ability of large multimodal models.

\subsection{State-Grounded Planning}

Based on the formulation in Section \ref{sec:formulation}, the objective of the embodied planning model $M$ is to generate a plan $P$ based on the textual user inputs $X_\text{req}$ and visual environment inputs $X_\text{img}$. For exact matching during evaluation, the model is also provided with the location names $X_\text{loc}$ and object names $X_\text{obj}$. 
Hence, the general objective can be formulated as $\hat{P} = M(X_\text{req}, X_\text{img}, X_\text{loc}, X_\text{obj})$, where $\hat{P} = \{(a_1, o_1), \ldots, (a_l, o_l)\}$ is the generated action plan, with $a$ denoting the action type and $o$ denoting the corresponding location or object of each action.
However, this objective poses a challenge in that it does not explicitly consider the initial environment state $S_\text{init}$ or the goal state $S_\text{goal}$, where the states represent the mapping between the specific locations and objects.
Concretely, the model requires visual perception of the environment to determine the initial state, and comprehension of the user request to interpret the goal state.
Without first grounding the planning process in the environment states, the model may output an inaccurate plan.

Therefore, our framework as shown in Figure \ref{fig:framework} introduces state-grounded planning, which explicitly generates and conditions on the initial environment state and goal state before starting to generate the action plan. 
Concretely, we need to modify the planning objective to first consider the initial state $\hat{S}_\text{init}$ and goal state $\hat{S}_\text{goal}$. As the model does not have access to the ground-truth initial state or goal state in practice, we require the model to self-generate the environment states in a sequential manner.
Note that the action types are a predefined list, included in user inputs.
Thus, we first consider the visual perception stage by explicitly generating the initial state $\hat{S}_\text{init}$ based on the inputs:

\begin{align}
    \hat{S}_\text{init} = M(X_\text{req}, X_\text{img}, X_\text{loc}, X_\text{obj})
\end{align}

Consequently, we explicitly consider the goal comprehension stage by generating the goal state $\hat{S}_\text{goal}$ based on the multimodal inputs and initial state $\hat{S}_\text{init}$:

\begin{align}
    \hat{S}_\text{goal} = M(X_\text{req}, X_\text{img}, X_\text{loc}, X_\text{obj}, \hat{S}_\text{init})
\end{align}

Finally, we are able to ground the reasoning for plan generation with the previously generated environment states $\hat{S}_\text{init}$ and $\hat{S}_\text{goal}$ in order to generate the action plan $\hat{P}$:

\begin{align}
    \hat{P} = M(X_\text{req}, X_\text{img}, X_\text{loc}, X_\text{obj},  \hat{S}_\text{init}, \hat{S}_\text{goal})
\end{align}

In practice, as the outputs are autoregressively generated by the large multimodal models, we can unify the three stages of the grounded planning process in a seamless and sequential manner, with each stage depending on the outputs of the previous stage. 
This ensures that the initial state, goal state, and action plan are coherently generated in a single continuous process by the same model $M$, with minimal additional cost.
Specifically, we provide demonstration examples as mentioned in Section \ref{sec:setting}, which guide the model to generate the environment states and action plan in a seamless process.

\begin{figure*}[t!]
\centering
\resizebox{0.8\textwidth}{!}{
\begin{tikzpicture}
    \definecolor{customRed_1}{HTML}{EEBEC0}
    \definecolor{customRed_2}{HTML}{F5867F}
    \definecolor{customRed_2_5}{HTML}{E73C36}
    \definecolor{customRed_3}{HTML}{C94E65}
    \definecolor{customYellow_1}{HTML}{EFDBB9}
    \definecolor{customYellow_2}{HTML}{FFC839}
    \definecolor{customYellow_2_5}{HTML}{F1B656}
    \definecolor{customYellow_3}{HTML}{D9995B}
    \definecolor{customBlue_1}{HTML}{A3CDEA}
    \definecolor{customBlue_2}{HTML}{63ADEE}
    \definecolor{customBlue_2_5}{HTML}{397FC7}
    \definecolor{customBlue_3}{HTML}{015493}
    \begin{axis}[
        width=11cm,
        height=5.8cm,
        ybar=0cm,
        bar width=6pt,
        ymax=100,
        ymin=20,
        xmax=3.3,  
        xtick = {1,2,3,4,5},
        xticklabels = {Claude Opus, Gemini Pro, GPT-4V},
        xtick pos = left,
        ytick pos = left,
        ymajorgrids = true,
        ytick={0,20,40,60,80,100},
        enlarge x limits=0.22,
        grid style=dashed, 
        legend style={
            legend style={row sep=-0.1cm},
            at={(1.6,1.002)},
        },
        legend image code/.code={
          \draw[#1] (0cm,-0.1cm) rectangle (0.4cm,0.05cm);
        }, 
        legend cell align={left},
    ]
    \addplot[fill=customRed_1] coordinates {(1, 31) (2, 47) (3, 37)};
    \addlegendentry{Commonsense: Baseline};
    \addplot[fill=customRed_2] coordinates {(1, 30) (2, 47) (3, 36)};
    \addlegendentry{Commonsense: Chain-of-Thought};
    \addplot[fill=customRed_2_5] coordinates {(1, 28.3) (2, 43.9) (3, 41.4)};
    \addlegendentry{Commonsense: Reasoning Via Planning};
    \addplot[fill=customRed_3] coordinates {(1, 36) (2, 49) (3, 39)};
    \addlegendentry{Commonsense: \methodname{} (Ours)};
    \addplot[fill=customYellow_1] coordinates {(1, 64) (2, 78) (3, 77)};
    \addlegendentry{Physical: Baseline};
    \addplot[fill=customYellow_2] coordinates {(1, 63) (2, 75) (3, 79)};
    \addlegendentry{Physical: Chain-of-Thought};
    \addplot[fill=customYellow_2_5] coordinates {(1, 61.0) (2, 79.0) (3, 79.1)};
    \addlegendentry{Physical: Reasoning Via Planning};
    \addplot[fill=customYellow_3] coordinates {(1, 70) (2, 83) (3, 84)};
    \addlegendentry{Physical: \methodname{} (Ours)};
    \addplot[fill=customBlue_1] coordinates {(1, 48) (2, 44) (3, 43)};
    \addlegendentry{Safety: Baseline};
    \addplot[fill=customBlue_2] coordinates {(1, 48) (2, 46) (3, 48)};
    \addlegendentry{Safety: Chain-of-Thought};
    \addplot[fill=customBlue_2_5] coordinates {(1, 48) (2, 46) (3, 48)};
    \addlegendentry{Safety: Reasoning Via Planning};
    \addplot[fill=customBlue_3] coordinates {(1, 47.8) (2, 47.8) (3, 55.0)};
    \addlegendentry{Safety: \methodname{} (Ours)};
    \end{axis}
\end{tikzpicture}
}
\vspace{-0.6em}
\caption{Main results of plan validity (\%) across different embodied planning strategies. We report average performance in each task category.}
 \vspace{-1em}
\label{fig:main_results}
\end{figure*}
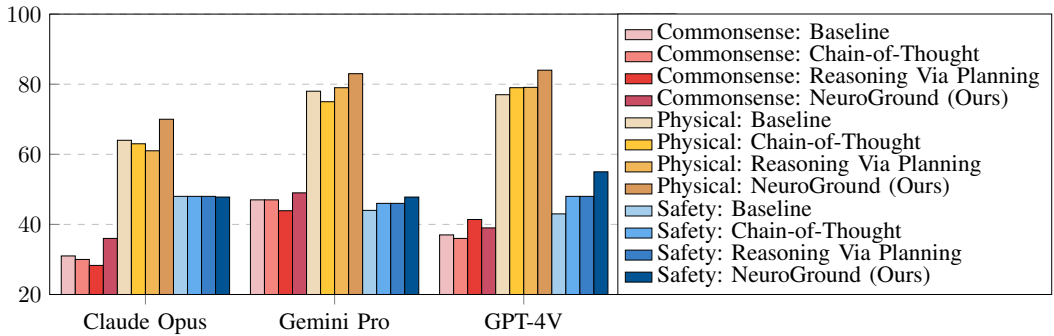

\subsection{Neuro-Symbolic Augmentation}

To further address the reasoning challenges for plan generation as discussed in Section \ref{sec:bottlenecks}, we augment the planning process with a symbolic solver.
For example, as shown in Figure \ref{fig:framework}, the model $M$ may generate a correct initial state $\hat{S}_\text{init}$ and goal state $\hat{S}_\text{goal}$, but still generate an invalid plan due to hallucination.
In this case, the plan attempted to pick the durian object from the pan location, which contradicts the initial state.
On the other hand, many well-established planning algorithms can search for the optimal and valid plan when given the initial and goal states \cite{plangoals, shirai2024vilain}.

Thus, we incorporate a symbolic planning engine based on the A* search algorithm \cite{astarsearch} in the unified-planning library\footnote{\url{https://unified-planning.readthedocs.io/en}}.
Concretely, given the generated initial state $\hat{S}_\text{init}$ and goal state $\hat{S}_\text{goal}$, we directly parse and input the states into the planning engine $E$ to determine the action plan:

\begin{align}
    \hat{P}_\text{engine} = E(\hat{S}_\text{init}, \hat{S}_\text{goal})
\end{align}

However, we observe challenges in directly using the model-generated states with the planning engine.
Notably, there may be cases of parsing issues when the model does not follow the structured state format.
Thus, we leverage the engine plan when possible, but otherwise fall back to the model-generated plan for robustness. 

\section{Experiments}


To evaluate our framework for embodied planning, we use the same task formulation and evaluation setting as detailed in Section \ref{sec:formulation} and Section \ref{sec:setting}. 
Notably, the model inputs contain multimodal content such as visual images depicting the environment and natural language user instructions.
The main output will be the action plan, specifying that sequence of actions that the agent should perform.
Our main evaluation metric is the average plan validity, based on whether the plan satisfies the goal state constraints.

\subsection{Evaluated Large Multimodal Models}
\label{sec:models}

To investigate the limits of embodied planning abilities, we focus our study on the state-of-the-art large multimodal models at the time of writing.
Specifically, we consider the following three models which are widely popular and show leading performance on multimodal tasks \cite{yue2023mmmu}:
1. Claude Opus: We select the highest-performing model from Anthropic's Claude model family \footnote{\url{https://www.anthropic.com/claude}}, which emphasizes near-human levels of comprehension and fluency on complex tasks and sophisticated vision capabilities on par with other leading models. Specifically, we use the ``claude-3-opus-20240229'' API model version. 
2. Gemini Pro: We include the most recent model from Google's Gemini model family\footnote{\url{https://deepmind.google/technologies/gemini/pro/}}, a highly capable multimodal mixture-of-experts model capable of recalling and reasoning over fine-grained information. We use the ``gemini-1.5-pro-latest'' API model version, which is the latest and most capable model in the Gemini model family at the time of writing.
3.  GPT-4V: We select the premier \cite{sparks} multimodal model of OpenAI's GPT-4 model family\footnote{\url{https://openai.com/index/gpt-4v-system-card/}}, which enables users to instruct the model to analyze text and image inputs. Specifically, we use the ``gpt-4-vision-preview'' API version.

\subsection{Planning Methods}

To demonstrate the effectiveness of our planning framework, we compare it to existing planning methods.
We note that while there are several related works in planning \cite{yao2023tree}, they may be task-specific and not applicable to our embodied planning task in the multimodal setting. 
For example, Inner Monologue \cite{huang2022innermonologue} requires a live environment which is difficult to replicate, while ViLaIn \cite{shirai2024vilain} requires bounding box inputs that are not applicable to our setting.
Hence, we restrict to general planning approaches that are easily applicable to our task setting as formulated in Section \ref{sec:formulation}:
1. Direct Planning Baseline: Following the baseline planning method in the preliminary study of Section \ref{sec:bottlenecks}, the model is prompted to generate the action plan directly, based on the multimodal inputs.
2. Chain-of-Thought: We compare against chain-of-thought prompting as a planning approach, as it is widely applicable to many reasoning and problem-solving tasks, including embodied planning \cite{vila}. We follow the planning format shown in the original work \cite{wei2023chainofthought}.
3. Reasoning Via Planning (RAP): We also include this comparison approach which uses the model as a reasoning agent for planning \cite{hao2023reasoningrap}. 


\subsection{Evaluation Results}

We report the main evaluation results of our planning framework in Figure \ref{fig:main_results}.
Compared to the other methods, our framework shows notable and relatively consistent planning improvements across different categories and state-of-the-art models.
We observe that chain-of-thought prompting which leverages intermediate reasoning steps does not consistently improve over the baseline approach of directly generating the plan.
This may indicate reasoning steps alone are insufficient to ground the model, compared to our approach which explicitly generates and considers the initial and goal states of the environment before planning.
Furthermore, reasoning via planning also shows inconsistent results despite simulating the future states of each action in the plan. 
On the other hand, our framework does not need to repeatedly simulate the future state, and directly considers the generated initial state and goal state. 
Thus, it can naturally condition the model to generate more accurate plans.

\paragraph{Ablation}
To evaluate different components of our framework, we include the ablation study in Table \ref{tab:ablation}.
Notably, we find that not using the symbolic engine component results in less accurate plans, indicating that the neuro-symbolic augmentation can mitigate some model-generated plan failures as shown in Figure \ref{fig:framework}.
Furthermore, we observe that even without the neuro-symbolic augmentation, the performance is greater than the baseline and chain-of-thought prompting, which indicates that the state-grounded planning component of our framework confers significant benefits.



\begin{table}[!t]
\centering
\caption{Ablation study with average scores.}
\vspace{-0.6em}
\resizebox{\linewidth}{!}{
\begin{tabular}{lccccc}
\toprule
 & {Claude} & {Gemini} & {GPT-4V} \\
 \midrule
Baseline & 45.3 & 56.6 & 52.1 \\
Chain-of-Thought Prompting & 45.7 & 57.3 & 54.2 \\
Reasoning Via Planning (RAP) & 43.7 & 58.7 & 55.4 \\
\midrule
\methodname{} & 50.8 & 61.6 & 57.8 \\
\quad w/o Symbolic Engine & 50.5 & 58.1 & 56.5 \\
\bottomrule
\end{tabular}
}
\vspace{-1em}
\label{tab:ablation}
\end{table}

\section{Conclusion}

In this work, our research addresses the urgent need to advance large multimodal models in the area of embodied planning. We introduce \datasetname{}, a novel dataset designed to benchmark these models in diverse and complex scenarios, pushing beyond the limitations of existing benchmarks. Through our fine-grained analysis, we identify significant bottlenecks in current state-of-the-art models, particularly in visual perception, comprehension, and reasoning. To mitigate these challenges, we propose \methodname{}, a neuro-symbolic framework that enhances model-generated plans by first grounding them in the environment states and then augmenting them with symbolic engines. Our experimental results demonstrate consistent benefits over existing methods, underscoring its potential to significantly advance the embodied planning capabilities of large multimodal models.

\section*{Acknowledgement}

This work was substantially supported by DAMO Academy through DAMO Academy Research Intern Program. 

\bibliographystyle{IEEEtran}
\bibliography{custom}

\end{document}